\documentclass[conference]{IEEEtran}
\IEEEoverridecommandlockouts
\pdfoutput=1
\usepackage{cite}
\usepackage{amsmath,amssymb,amsfonts}
\usepackage{algorithm}  
\usepackage{algorithmicx}  
\usepackage{algpseudocode}

\usepackage{graphicx}
\usepackage{textcomp}
\usepackage{xcolor}
\usepackage{subfigure}
\usepackage{url}

\algrenewcommand{\algorithmiccomment}[1]{\hskip0em $\triangleright $ #1}

\def\BibTeX{{\rm B\kern-.05em{\sc i\kern-.025em b}\kern-.08em
    T\kern-.1667em\lower.7ex\hbox{E}\kern-.125emX}}
\begin{document}

\title{Towards Generalization and Data Efficient Learning of Deep Robotic Grasping}

\author{Zhixin Chen, Mengxiang Lin*, Zhixin Jia, Shibo Jian \\
\IEEEauthorblockA{\textit{State Key Laboratory of Software Development Environment} \\
\textit{School of Mechanical Engineering and Automation, Beihang University}\\
Beijing, China \\
\{zhixinc, linmx, jiazx, jianshibo\}@buaa.edu.cn}
\thanks {* Corresponding author}
}

\maketitle

\begin{abstract}
Deep reinforcement learning (DRL) has been proven to be a powerful paradigm for learning complex control policy autonomously. 
	Numerous recent applications of DRL in robotic grasping have successfully trained DRL robotic agents end-to-end,
	 mapping visual inputs into control instructions directly, but the amount of training data required may hinder these applications in practice. 
	 In this paper, we propose a DRL based robotic visual grasping framework, in which visual perception and control policy are trained separately rather than end-to-end. 
	 The visual perception produces physical descriptions of grasped objects and the policy takes use of them to decide optimal actions based on DRL. 
	 Benefiting from the explicit representation of objects, the policy is expected to be endowed with more generalization power over new objects and environments. 
	 In addition, the policy can be trained in simulation and transferred in real robotic system without any further training.
	  We evaluate our framework in a real world robotic system on a number of robotic grasping tasks, such as semantic grasping, clustered object grasping, moving object grasping. 
	  The results show impressive robustness and generalization of our system.

\end{abstract}


\section{Introduction}

\label{intro}
As one of the most important manipulation skills, vision-based grasping has been studied intensely in the field of robotics for decades.
Although robots performing pick and place tasks have been applied successfully in industry, creating autonomous robots for grasping objects in unstructured real-world scenes remains an open question.
In recent years, deep reinforcement learning (DRL) has attracted increasing research attention in robotics since its success in video game.
Combining with CNNs, DRL maps the features of visual observations directly into control policies by trial-and-error.
This provides a general way for robots to learn manipulation skills by using information acquired from cameras \cite{simulation-robotic}, \cite{dexterous-manipulation}.

The typical way to train a visual-based DRL agent is in an end-to-end fashion, in which the reward signal of reinforcement learning is used to train both CNNs and policy networks synchronously.
However, in order to achieve satisfying performance, large amounts of interaction data are required for training of a DRL agent.
For example, to collect enough training data, \cite{hand-eye} executed 800,000 robotic grasping attempts in several months with 14 robots,
and \cite{self-super} collected 700 hours robot grasping data with 50,000 grasping attempts.
Moreover, DRL methods try to map raw visual observation into a lower dimensional latent space of control that preserves various types of information about manipulated objects.
However, tangled and uninterpretable latent representation restricts the generalization across object and environment and further leads to poor control policy.
Most works have to evaluate the trained DRL agents with the similar objects and environments as that of training \cite{robotic-autoencoder}, \cite{dexterous-manipulation}, since the networks trained before have to be fine tuned to adapt to change and hundreds of thousands of robotic manipulation experiences may be needed once more when transferring to a new environment.
These limitations will definitely prohibit the use of DRL method in real-world robotic application. 

A feasible way to alleviate data requirements is to train DRL agents in simulation, in which the interaction could be speed up easily by using programming techniques, such as multithreading. 
With this approach, large volumes of experiences can be captured efficiently than that of real world interactions and meanwhile many variants of environments could be constructed for generalization concern.
However, there is a huge gap between the simulation and the real-world, which causes the agents trained in simulation to be hardly applied in real world conditions, especially in the context of vision-based robotic manipulation, where illumination changes and varying textures can have significant effects on the quality of the results.
Therefore, some DRL based robotic control approaches are only verified in simulation due to difficulties in transferring from simulation to real robots \cite{no-real-robot2}, \cite{no-real-robot1}.
To alleviate the problem some techniques are proposed to allow for automatic adaptation to the real world environments \cite{simulation2real2}, \cite{simulation2real}.

In this paper, we propose a DRL-based visual grasping system aiming at improving generalization performance with the least cost of the acquisition of real world experiences. 
Following the typical visual based DRL paradigm, our framework consists of two major components: a CNN based visual \emph{perception} and a DRL based control \emph{policy}.
The perception module extracts features from visual observation (i.e. raw images) and then the features are mapped into the action space by the policy module.

We train the perception and the policy separately instead of end-to-end.
The perception is trained in a supervised setting to produce the semantic and spatial information of the grasped objects.
 In the meantime, the control policy is trained in a simulation environment where the class and pose of the object to be grasped can be read automatically.
 Training the policy with the quantitative description of manipulated objects can be beneficial to both generalization and transferability since the information irrelevant for control decision is discarded.
 In our work, after roughly \textbf{30 minutes} of training in simulation, the policy is directly transferred to a real robotic system \textbf{without any additional training}.
 The performance of our system is evaluated on challenging tasks including semantic grasping, clustered object and moving object grasping.
 The experimental results demonstrate the robustness and generalization of our approach.

\section{Related work}
\label{relatedwork}
Simulation environments can provide experiences data much more effective 
since the simulation could be accelerated by programming. 
And many robotic manipulation DRL algorithms are verified in simulation environments \cite{simulation-robotic}, \cite{model-base}.
Unfortunately, the gap between simulations and real-world makes the the agent trained in simulation can hardly use in physical robot. 
Many works had tried to bridge reality gap \cite{simulation2real}, \cite{simulation2real2}.
In \cite{simulation2real}, the images came from simulations were rendered in randomization, and while the visual perception had seen enough variability over simulations, the real-world images may appear to the model just as another variation.
Such randomization made a successful visual grasping model in real-word.
\cite{simulation2real2} unified the visual inputs from simulation and real-world using an adaptation network.
The adaptation network was trained to generalize canonical images from randomized images from simulation.
And because of the randomization, the trained network could also generalize canonical images from real-world.
The generalized canonical images which had been mapped into the same space were used to train the visual grasping DRL agents.
Benefited from the adaptation network, DRL agents could be trained in simulation and used in real-world.

Many researches have tried to relieve data inefficiency by improving the efficiency of DRL training process and experiences data generation.
Guided policy search (GPS) algorithm \cite{guided-policy-search} converts reinforcement learning into supervised learning,
where a trained a local linear controller provided with full state observation (i.g., object poses) served as supervisor.
And a global policy parameterized by neural networks derived from supervision. 
This allows a CNNs policy with 92,000 parameters to be trained in tens of minutes of real-world robotic manipulation time 
and in test stage the full state is no longer available that the policy trained in a supervised setting could handle several novel, unknown configurations.
Another direction to improve sample efficiency is to accelerate model-free reinforcement learning with a learned dynamics models \cite{model-base}.
The learned models can generate synthetic sample data to enrich the agent experiences efficiently 
which has no need to execute the physical robot, though it needs additional efficient model learning algorithms \cite{model-learn}, \cite{model-learn-3}.
However, the learned model would quite differ from the true dynamics and the induced error would weaken performance of learned policy.

These methods tried to train an optimal policy and visual perception simultaneously in an end-to-end style.
However, it can hardly be generalized to different manipulated objects and different execution environments.
Since the generalization is relied on the distribution of training data,
it requires a huge experience data to achieve usable generalization ability \cite{e2eGeneralize}.
It is impractical in a robotic grasping task to acquire enough data.
An intuitive alternative is to train image representation and reinforcement agent separately \cite{auto-encoder}, \cite{robotic-autoencoder}.
With an auto encoder pretrained by an auxiliary reconstruct loss, the high dimension of image input is embedded into a low dimension, latent space and aggregate useful features before interacting with environment. 
This way the training of the reinforcement agent networks would be more easily with much less interaction experiences 
for there is no need to learn the state representation and the training would significantly speed up. 
However, the latent feature representation has no exact physical meanings and would be lack of interpretability as well the trained policy. 
From this perspective, meaningful feature representation would significantly improve generalization ability.

\section{Framework}
\label{framework}

We propose a robotic grasping framework based on deep reinforcement learning. 
Reinforcement learning enables agents (e.g., robots) to learn an optimal policy through interaction with environments by trial-and-error. 
In doing so, we formulate a robotic grasping problem as a Markov decision process: 
at time $t$, the robot receives the state of target objects and constructs environment state $s_t$ accordingly. 
After that, the robot chooses an action $a_t$  to move itself based on the current policy $\pi \left(a_{t}|s_{t}\right)$. 
Then the environment transits to a new state $s_{t+1}$ reacting to $a_t$ and an immediate reward $R_t \left(s_t,a_t,s_{t+1}\right)$ is offered by the environment. The goal of the robot is to find an optimal policy  $\pi ^{*}\left(a_{t}|s_{t}\right)$ that maximizes the discounted future rewards

$$G_t=\sum _{i=t}^{\infty}\gamma ^{i-t}R_{i}\left(s_i,a_i,s_{i+1}\right)$$ 
where  $0 < \gamma < 1$ is the discounted factor which reduces the weight of future rewards.

\begin{figure*}
	  \includegraphics[width=1\textwidth]{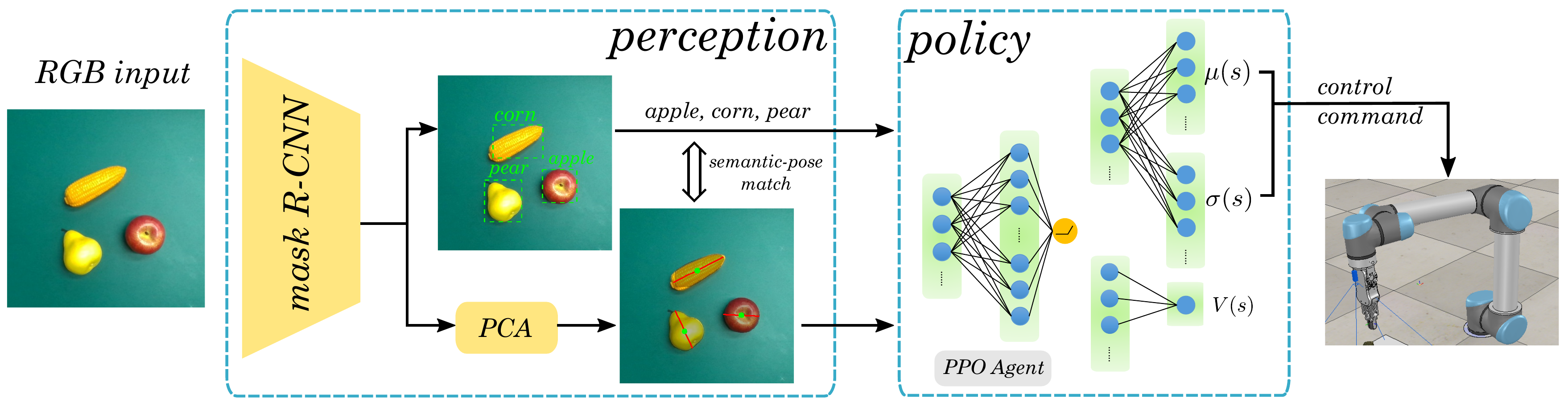}
	\caption{Our framework for robotic prehensile control.} 
	\label{framework_figure}       % Give a unique label
\end{figure*}

Similar to recent works  \cite{robotic-autoencoder}, \cite{guided-policy-search}, our framework is composed of two stages, as shown in Fig.\ref{framework_figure}. 
Raw RGB images from camera are input into the perception, where object semantic segmentation and pose estimation are made by Mask R-CNN and PCA respectively. 
The policy is a PPO \cite{ppo} agent which receives the control quantities of desired objects and decides which action will be taken to execute grasping. 
The pseudo code of grasping a single object with our framework is presented in Algorithm \ref{framework_pseudo}. 
The details of each component are discussed in the following subsection.

The perception and policy are trained separately. 
In particular, Mask R-CNN is trained in a supervised way, in which the labels are constructed manually using a tool LabelMe \cite{labelme}. 
There is no training stage for PCA as it is an unsupervised method. 
The PPO is trained in simulation for fast experience data acquisition. 
Since the semantic and position information can be read by an interface provided by the simulation environment, the training of PPO could proceed in parallel with that of Mask R-CNN. 

\begin{algorithm}
	\caption{Separate Perception and Policy}
	\label{framework_pseudo}
	\begin{algorithmic}[1]
		
		\Require $image$
		\Statex
		\Statex \textbf{\emph{Perception}}
		\State \Comment detect object class and mask from raw images
		\State  $mask, class$ $\gets$ \textbf{mask-rcnn}($image$)
		\State \Comment get physical quantities from mask
		\State $center, direction$ $\gets$ \textbf{PCA}($mask$)
		\Statex
		\Statex \textbf{\emph{Policy}}
		\Repeat
			\State \Comment decide action via PPO
			\State $action$ $\gets$ \textbf{PPO}($center, direction$)
		\Until{$|action| < \epsilon$}
		\Statex
		\State \Comment grasp specific object and success check
		\State $success$ $\gets$ \textbf{grasp}($class$)
		
	\end{algorithmic}
\end{algorithm}

\subsection{Perception}
\label{perception}

The perception plays a sensor-like role that transforms raw image inputs into physical quantities binding with object semantic information (i.e., object class and its corresponding pose). 
We should note that this work focuses on a 3DOF grasp \cite{grasp-pose} given that the workspace of a robot is constrained to a tabletop. 

\subsubsection{Semantic Segmentation} 

To grasp a target object, the robot must know where the target is. 
To achieve this, we leverage a popular semantic segmentation method Mask R-CNN \cite{maskrcnn} as the front part of the perception to detect and segment objects from raw images. 

Based on Faster R-CNN \cite{fastrcnn}, Mask R-CNN introduces a mask branch at the end of the original network for segmentation tasks.
It proceeds in two stages: first, the region proposal network (RPN) \cite{fasterrcnn} is applied to scan the image to find the area where the target exists; 
secondly, the object class and its bounding box coordinates are predicted simultaneously. 
Finally, a pixel-wise binary mask for each bounding box is generated by a fully convolutional network (FCN) indicating whether each pixel in bounding box is the point of the detected object. 
As a result, masks on the original image exactly cover the areas where the objects exist. 
The class and mask of an object provide us with a good starting point for pose estimation.

\subsubsection{Object Pose}

Since an object mask produced carries information about object pose, 
learning a DRL policy from masks is in principle possible, as what most DRL based approaches do. 
In realistic robotic applications however, we can not afford to collect such huge interaction data required by a policy learning algorithm. 
To avoid this difficulty, we further infer pose for an object instance based on the object mask obtained. 
In a 3DOF grasp setting, object pose can be represented by a 2-dimensional position coordinates of the object center and the direction of the object. 
Here, we develop a Principal Component Analysis (PCA) based method to estimate 3D object pose from a pixel-wise binary mask output by Mask R-CNN. 
In general, PCA is an unsupervised method that could identify the main components of data with largest variances from a big dataset. 
For our purpose, the center and main direction of a set of pixel points are inferred by using PCA.

The output of Mask R-CNN is an object with its covered mask which contains all pixel points formalized as 
$$mask=\left \{ \left(x_0,y_0\right), \left(x_1,y_1\right),...,\left(x_n,y_n\right)\right \} $$
where $n$ is the number of pixel points in the mask, i.e the number of samples in PCA.
Firstly, we calculate the mean point of $mask$ as the center point of the mask:
$$
c=(x,y)=\frac{1}{n} \sum_{i=1}^n (x_i, y_i)
$$ 
Note that, the mean point $c=(x,y)$ is the geometric center of a mask.  
After that, all the points in the $mask$ are subtracted by the $c$ resulting residual coordinates
$$Res = mask - c$$
Thus, the covariance matrix of $Res$ and its corresponding eigenvalues and eigenvectors are calculated:
$$
\lambda _1, \lambda _2 = eigenvalues \left ( covMat\left(Res\right) \right )
$$
$$
\alpha _1, \alpha _2 = eigenvectors \left ( covMat\left(Res\right) \right )
$$ 
Since the pixel points are in two dimensions, there are totally two eigenvalues and eigenvectors of $Res$ matrix. 
The $\lambda$ and $\alpha \in \mathbb{R}^{2\times 1}$ are sorted by the magnitude of eigenvalues in descending order. 
Finally, the main component with largest variance is calculated: 
$$
M = Res \cdot \alpha _1 \cdot \alpha _1^\top + c
$$
$M \in \mathbb{R}^{n\times 2}$  contains $n$ points on a straight line. 
We take two points from $M$ randomly to construct a straight line. 
$\theta$  represents the angle of the straight line respect to the horizontal axis. 

\begin{figure*} %figure环境，可于此处加[h]来固定位置。
	\centering  %使图片居中显示
	\includegraphics[scale=1]{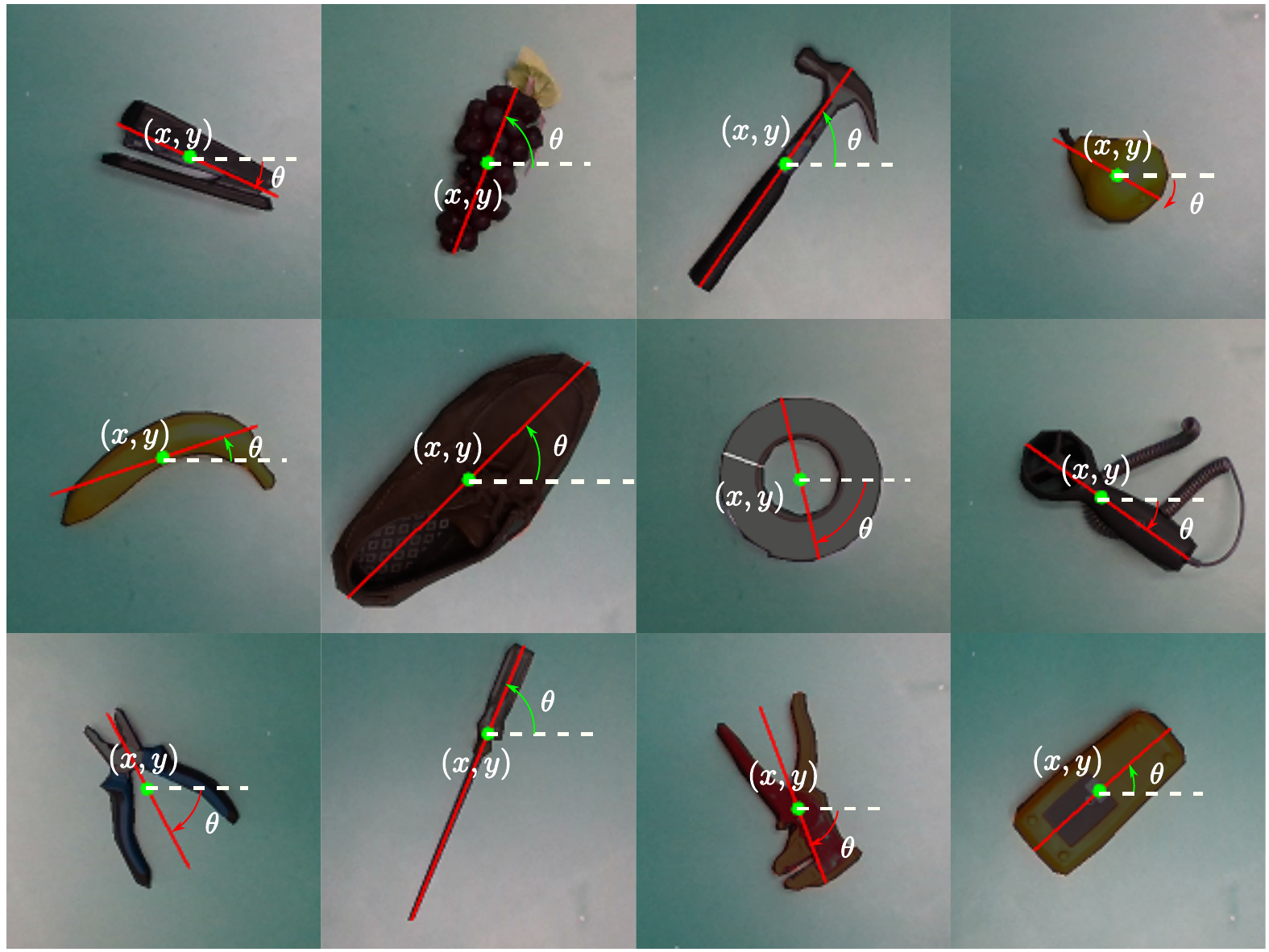} %图片尺寸及位置。
	\caption{
	The example results of PCA pose estimation. 
	The black shadows represent the masks produced by Mask R-CNN, and the green points are the center points of the masks, i.e. 
	the position coordinate of the object, and the red lines indicate the main component of the masks, 
	i.e. the orientation of the object in the plane. 
	The range of $\theta$ is $\left [-\frac{\pi}{2}, \frac{\pi}{2}\right ]$, where the green arrow indicates a positive $\theta$, while the red arrow is negative. } %图题
	\label{PCA-result} %标签，方便在文章中引用图片。
\end{figure*}

Fig.\ref{PCA-result} shows the results of PCA on a number of objects with various shapes. 
With the help of a calibrated camera, the position and orientation in pixel coordinates can be mapped into that of a physical coordinate system.
 %figure环境

\subsection{Policy}
\label{policy}

The policy is a deep reinforcement agent that receives the physical quantities from the perception and decides an optimal action to move the robot. For our framework, 
we adopt a policy gradient method called Proximal Policy Optimization (PPO) \cite{ppo} which is favorable for high dimension continuous robotic control problem. 
PPO significantly improves data efficiency over other policy gradient methods by updating multiple gradient steps with the same trajectory. 
Moreover, to avoid an increase in data variance, PPO introduces a clipping coefficient which stops the gradient when the difference between the updated and original policy is too large. 

\subsubsection{State Representation} 
\label{state-representation}

We concatenate the output from the perception $x, y, \theta$ (the pose of a target object) and the robotic configuration $x_r, y_r, \theta _r$ ( the pose of the end effector of a robot) to form the environment state $s_{t}$ as the input of the policy network PPO:
$$
s_{t} = \left(x,y,\theta,x_r, y_r, \theta _r\right)
$$
Through several fully connected layers, PPO finally outputs an action distribution over current state $s_t$.
Thereafter, the optimal action in the current state $s_t$ at time step $t$ is sampled from the action distribution:
$$
a_t \sim \pi\left (a|s_t \right )
$$
The action is a three-dimensional continuous variable instructing the robot's next moving direction and magnitude. 
The execution of the action would lead to a new environment state $s_{t+1}$ and an immediate reward $R_{t}$ offered by environment.

\subsubsection{Reward} 
\label{reward}

The reward function for learning the policy $R_{t}$ is defined as: 
$$
R_{t} = \left\{\begin{matrix}
		-d_t-0.1 & away\\ 
		-d_t+0.1 & approaching\\ 
		1 & grasp~success\\
		-1 & grasp~failed\\
	\end{matrix}\right.
$$
where $d_t$ is the distance between the end effector's current and target position in time step $t$.

If $d_t$ is decreasing compared to the previous $d_{t-1}$, the end effector is $approaching$ the target and will receive a slightly positive reward addition, 
and otherwise, it is $away$ with a negative reward. 
In this way, we encourage the end effector to approach and trace the target object as soon as possible.
When the policy decides actions bounded in a very small magnitude for several time steps, 
the policy will decide to execute the grasping, i.e., the end effector moves down on $z$ coordinate and closes the gripper. 
A grasping is counted as a success if the gripper is not fully closed. 
With a larger reward for a successful grasping, the policy could learn the tracing target policy and grasping policy simultaneously.

\subsubsection{Training Loss} 
\label{traing-loss}

PPO is an Actor-Critic style \cite{a3c} algorithm and typically contains a value function and a policy function.
The value function $V_{s_t}$ which estimate the expected reward from a state $s_t$ is trained to minimize TD error \cite{rl-introduction}, whose loss function is defined as:
$$
L_{V} =  \left ( V\left(s_{t}\right)- \left (R_{t}+\gamma V\left(s_{t+1}\right)\right )\right )^{2}
$$
where $\gamma $ is the discounted factor.

The policy function $\pi(a_t|s_t)$ which decides an optimal action over a state $s_t$ is trained to maximize a novel surrogate objective \cite{ppo} according to the value function:
$$
L_{a} = \textbf{min}\left ( r_t A_t, clip\left ( r_t, 1-\epsilon, 1+\epsilon \right ) A_t \right )
$$
where $r_t = \frac{\pi \left (a_t|s_t \right )}{\pi_{old}\left (a_t|s_t \right )}$ is the importance sampling coefficient in which $\pi_{old}\left (a_t|s_t\right )$ is the behavior policy whose parameters are frozen during one update epoch. 
$A_t$ is the advantage function \cite{a3c} which indicates if the reward of current action is above average. 
And it could be estimated easily by $A_{t} =R_{t} + \gamma V\left(s_{t+1}\right)- V\left(s_{t}\right) $ or GAE method \cite{gae} according to the value function $V(s_t)$.
And a \emph{clip} is a function that limits the importance sampling value between $1-\varepsilon$ and $1+\varepsilon$ 
in order to avoid a large step update where $\varepsilon$ is the clipping coefficient which usually equals to $0.2$ \cite{ppo}. 

Therefore, the final loss function $L$ becomes
$$
L = L_V - L_a
$$
and the parameters of network are updated through gradient descent method according to $L$.

\section{Experimental Evaluation}
\label{experiments}
\subsection{Implementation}
\label{implementation}

To evaluate our approach, we implemented a visual-based grasping system based on the framework shown in  Fig.\ref{framework_figure}. 
The perception consists of a Mask R-CNN and a PCA procedure, which are executed in pipeline. 
The input images resized into $600\times 600$ are fed into the Mask R-CNN and a number of object instances covered with their masks are produced. 
For each mask produced, PCA is invoked to compute its position and orientation as the output of the perception. 
The implementation of the Mask R-CNN is based on \cite{maskrcnn_implementation}. 
Instead of using a pre-trained Mask R-CNN model on general objects datasets such as MSCOCO \cite{coco}, we train the Mask R-CNN on our own dataset considering detection accuracy. 
To this end, 1000 images of 21 classes of objects are collected and labeled with their mask ground truth manually by the label tool LabelMe \cite{labelme}.

\begin{figure}
	\centering  
	\includegraphics[scale=1]{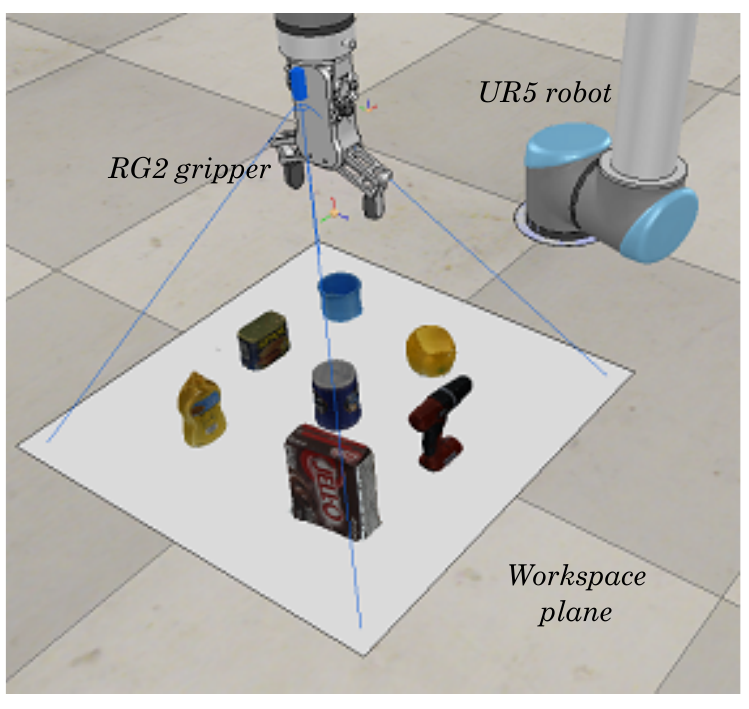} 
	\caption{Simulation environment set up in V-REP \cite{vrep}. }
	\label{simulation-environment} 
\end{figure}

For the policy, three fully-connected layers are stacked together to form a PPO agent. 
The first layer takes as input a six dimension vector concatenating the object position and the robot position and transforms the input into a 512 dimension latent vector.
And the second layer transforms the 512 dimension vector into two streams: a 512 dimension action vector and a 512 dimension value vector. 
Then one stream is transformed into two 3-dimension vectors, representing the parameters of action distribution $\mu$ and $\sigma$, while another stream is transformed into a scalar representing the value of current environment state. 
For efficient training of PPO, we setup a simulation environment in V-REP \cite{vrep}, as shown in Fig.\ref{simulation-environment}. 
Seven classes of objects from a robotic manipulation benchmarks YCB \cite{YCB} are used for PPO training in simulation, including a detergent bottle, an orange, a round can, a rectangular can, a cup, a pudding box and an electric drill. 
Since the class and pose of an object in the simulation environment can be obtained directly through software interfaces, PPO could be trained separately, without the help of the perception. 
The parameters of PPO are learned in a learning rate of $1e-5$ using Adam optimization method \cite{adam}.

\begin{figure}
	\centering 
	\includegraphics[scale=0.5]{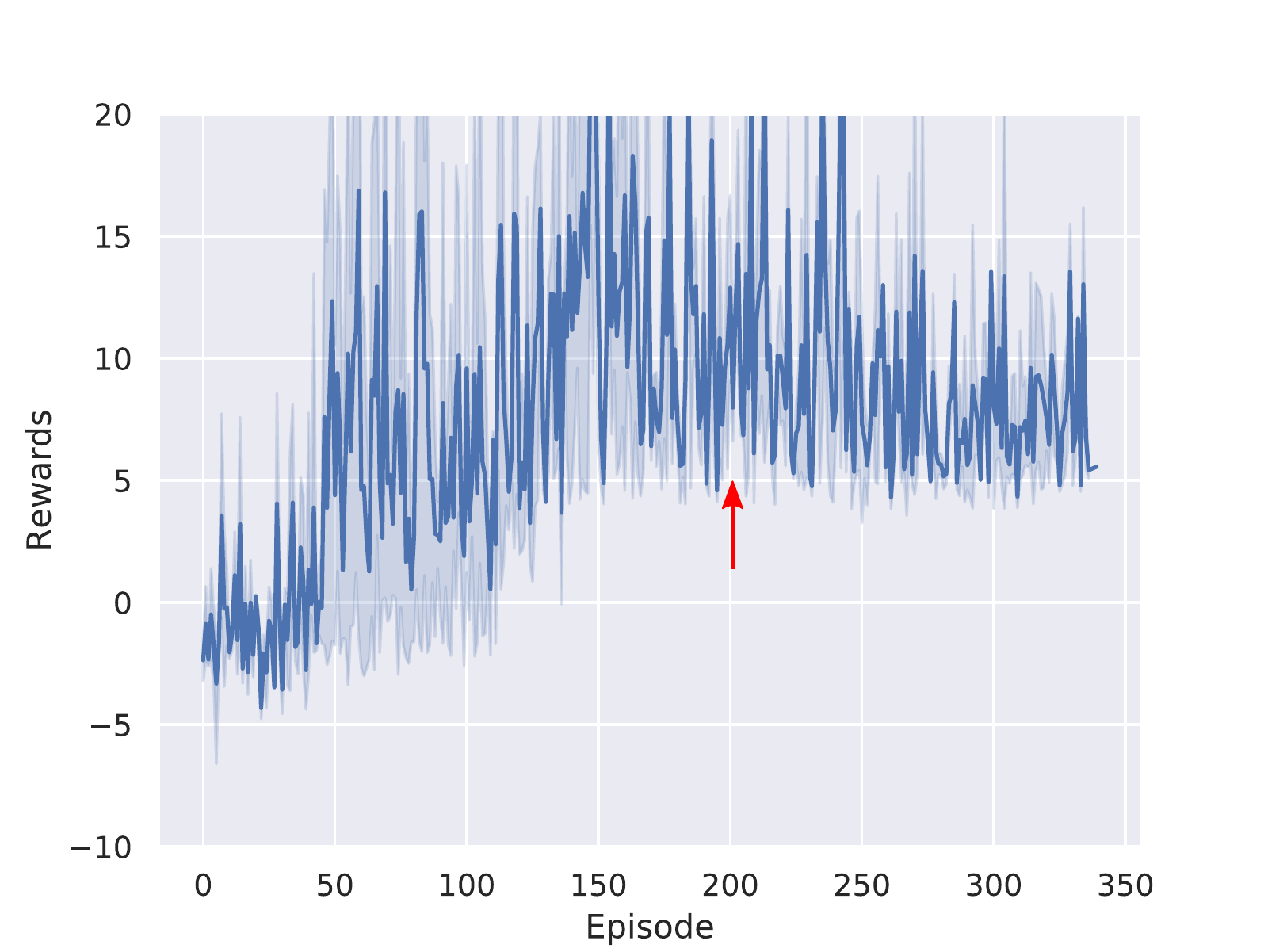} 
	\caption{The process of policy training in simulation. 
	The rewards converged very quickly and the model trained has achieved a good performance after about 30 minutes.
	 The policy model used in all experiments is trained for 200 episodes, as indicated by the red arrow.} 
	\label{train-result} 
\end{figure}

Both the training of PPO and Mask R-CNN is done on a PC with a RTX 2080Ti GPU. 
The average rewards of PPO training in simulation over 5 runs are shown in Fig.\ref{train-result}. 
Very impressive results are obtained after about \textbf{30 minutes} training of PPO, indicating by the red arrow in Fig.\ref{train-result}. 
By contrast, the interactions with the same number of episodes would take tens of hours for a real physical robot. 
The training of Mask R-CNN on our own dataset takes about 10 hours. 
This training may be not necessary since a pre-trained model on general dataset usually works well in many cases. 
It is worth noting that all the training above does not require a real-world robot and the trained networks will be transferred into a real-world robotic grasping system directly. 

\subsection{Real-world Evaluation} 
\label{real-world}

\begin{figure}
	\centering  
	\includegraphics[scale=1]{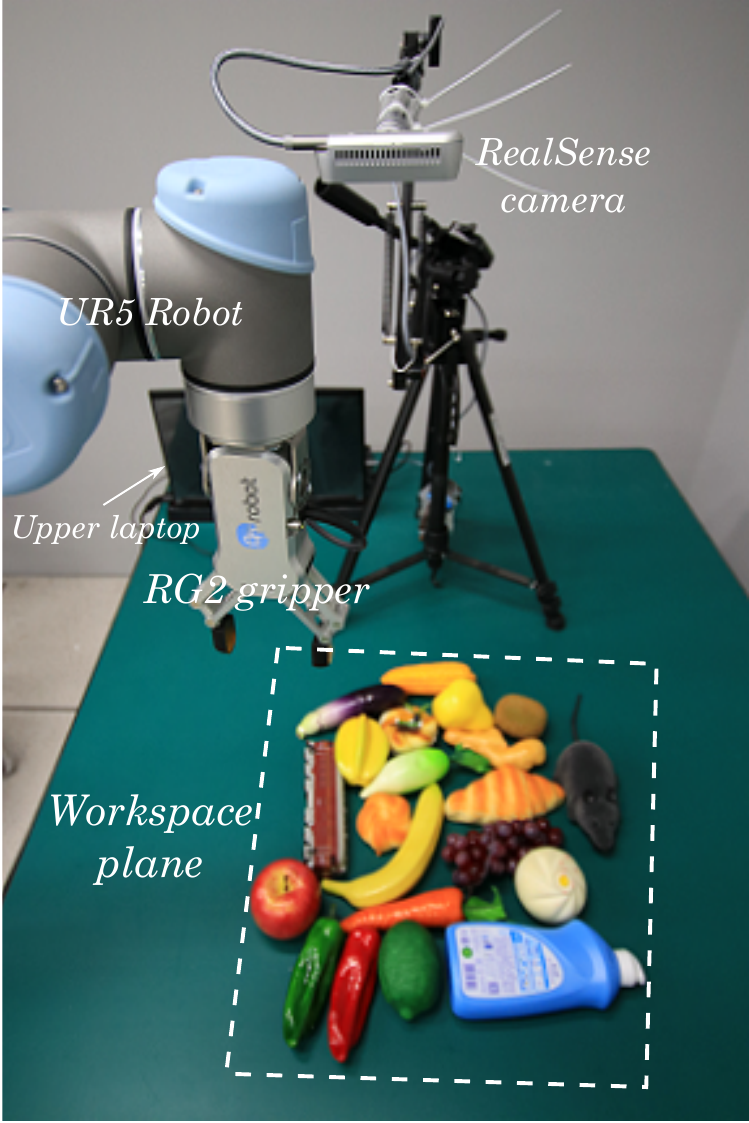} 
	\caption{The hardware setup of our system. }
	\label{real-world-environment} 
\end{figure}

The overall goal of our evaluation is to determine whether the trained networks can enable a real world robot to perform various grasping tasks without any further training. 
To this end, a number of grasp tasks commonly used in our daily life are designed to evaluate the ability to perform grasping skills and generalization over objects and situations.
We use an industrial UR5 robot arm with an RG2 gripper to achieve two-finger rigid grip. A RealSense camera \cite{realsense} is located 100 cm above the work surface, producing RGB images for input. 
A laptop with a RTX 2080 GPU acceleration is used for real-time robotic control and communication with UR5 via TCP/IP protocol. 
The experimental hardware platform is shown in Fig.\ref{real-world-environment}. 
It is worth note that the objects used in the experiments are totally different from that of PPO training in simulation. 

\subsubsection{Sim-to-Real Transfer}
\begin{figure} %figure环境，可于此处加[h]来固定位置。
	\centering  %使图片居中显示
	\subfigure[]{ % 填子图图题
		\includegraphics[scale=0.85]{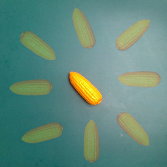}
	}
	\quad
	\subfigure[]{
		\includegraphics[scale=0.85]{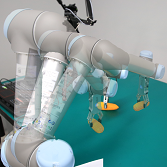}
	}
	\caption{Sim-to-Real transferring test of the trained policy.  
	(a) Picking up a corn in various position and orientation. 
	(b) Picking up a corn with different robotic configurations.} %图题
	\label{corn} %标签，方便在文章中引用图片。
\end{figure}

As mentioned before, the trained networks including Mask R-CNN and PPO are transferred into our robotic grasping system without any further training. 
We first examine the behavior of the system in a controlled manner. 
As shown in Fig.\ref{corn}, a target object (a corn) is placed on the work surface in various positions and orientations. 
The robot grasps the target successfully for 20 randomly chosen object locations.

\begin{figure} %figure环境，可于此处加[h]来固定位置。
	\centering  %使图片居中显示
	\subfigure[]{ % 填子图图题
		\includegraphics[scale=0.8]{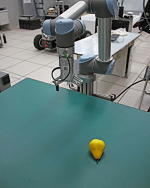}
	}
	\subfigure[]{
		\includegraphics[scale=0.8]{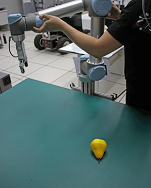}
	}
	\subfigure[]{
		\includegraphics[scale=0.8]{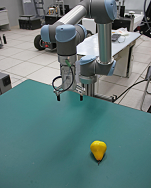}
	}
	\subfigure[]{
		\includegraphics[scale=0.8]{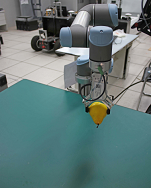}
	}
	\caption{Robustness test of the trained policy. 
	(a) The robot starts to perform a task. 
	(b) We forced it manually to an unseen configuration in training. 
	(c) The robot finds a proper path back to the right way. 
	(d) The robot grasps the target successfully.} %图题
	\label{trajectory} %标签，方便在文章中引用图片。
\end{figure}

Furthermore, in order to test the robustness of the control policy, we manually introduce external disturbances. 
As shown in Fig.\ref{trajectory}, the control policy could find its correct trajectory again and grasp the target successfully after a sudden change on the robot configuration during the robot's execution, exhibiting good stability and robustness.

\subsubsection{Multi-object Grasping}
\label{multiobj-grasp}

\begin{figure} %figure环境，可于此处加[h]来固定位置。
	\centering  %使图片居中显示
	\subfigure[]{ % 填子图图题
		\includegraphics[scale=0.90]{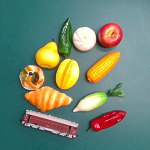}
	}
	\quad
	\subfigure[]{ % 填子图图题
		\includegraphics[scale=0.90]{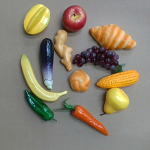}
	}
	\caption{Multiple object removing up. 
	(a) 11 objects in a green background. 
	(b) 12 objects in a brown background.} %图题
	\label{multiobj} %标签，方便在文章中引用图片。
\end{figure}

 Multi-object grasping is a common task used to measure the performance for a vision based robotic grasping system. 
 In our test setting, 10-13 objects are placed randomly on the table and the UR5 robot is requested to pick up all objects sequentially and then put them out of the workspace.
In addition, the background color of the work surface is shifted from white into brown or green. Two example test settings in different backgrounds are shown in Fig.\ref{multiobj}. 
A grasp is successful if an object is grasped and threw aside, while a remove completion means no objects are left on the table. 
We perform 10 tests for each background and grasp success rate and remove completion rate are presented in Table  \ref{multiobj-result}.

\begin{table}

	\caption{The results of multiple objects remove up. }
	\label{multiobj-result}       % Give a unique label
	\begin{tabular}{ccc}
	
	\hline\noalign{\smallskip}
	scenarios & grasp success & remove completion  \\
	
	\noalign{\smallskip}\hline\noalign{\smallskip}
	brown & 100\%(112/112) & 100\%(10/10)\\
	green & 100\%(120/120) & 100\%(10/10)\\
	dense & 93.7\%(104/111) & 80\%(8/10)\\
	\noalign{\smallskip}\hline
	\end{tabular}
\end{table}

\subsubsection{Clustered object Grasping}
\label{in-cluster}

\begin{figure} %figure环境，可于此处加[h]来固定位置。
	\centering  %使图片居中显示
	\subfigure[]{ % 填子图图题
		\includegraphics[scale=0.90]{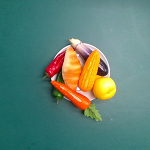}
	}
  \quad
  \subfigure[]{ % 填子图图题
		\includegraphics[scale=0.90]{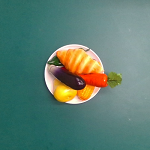}
  }
  \caption{Dense clustered object removing up.  
  (a) Objects blocked each other but every object is visible. 
  (b) The red pepper is completely invisible. } %图题
	\label{dense-cluster} %标签，方便在文章中引用图片。
\end{figure}

A challenge task in robotic manipulation is to grasp objects clustered closely together. 
As shown in Fig.\ref{dense-cluster}, in a cluster scenario, the objects would block each other and some objects may be completely invisible. 
In such a task, the order of manipulations really matters if we want to remove all the objects in sequence. 
To decide the ordering of picking up, we define a mask ratio $r$ for each object recognized as follows:
$$
r = \frac{m}{M}
$$
where $m$ is the recognized mask with possible occlusion and $M$ is the full mask of the object which is pre-determined. 
The larger the ratio, the more likely the object is to be picked up firstly as it is less occluded by others. 
For dense cluster scenarios, we perform 10 tests and grasp success rate and remove completion rate are presented in Table \ref{multiobj-result}. 
The failure cases occur due to misidentifications by Mask R-CNN because of partially visible objects.  

 %figure环境

\subsubsection{Semantic Grasping}
\label{semantic-grasp}

In a semantic grasping task, a robot is instructed to grasp a specified object among a set of candidates. 
The capability of semantic grasping is essential to allow autonomous robots to perform manipulations in an unstructured environment. 
Benefiting from the power of Mask R-CNN to detect objects, our system first identifies the class of an object before deciding how to pick up.
Similar to the experiment setting in multi-object grasping, 10-13 objects are randomly placed on the table for each trial. For each run, we randomly specify one object and simply count the number of successful grasps. 
We perform five trails and the success rate achieves \textbf{100\% (60/60)}.

\begin{figure*} %figure环境，可于此处加[h]来固定位置。
	\centering  %使图片居中显示
	\subfigure[]{ % 填子图图题
		\includegraphics[scale=0.35]{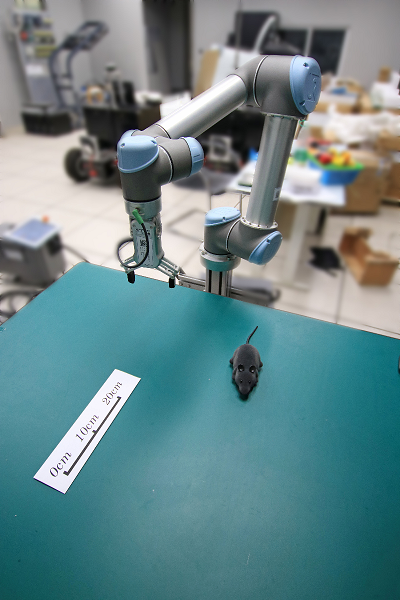}
	}
	\quad
	\subfigure[]{
		\includegraphics[scale=0.35]{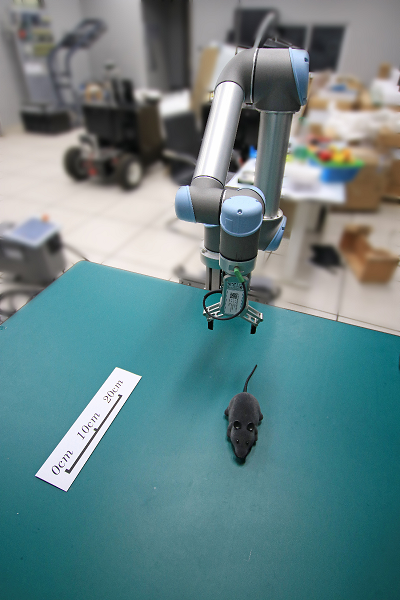}
	}
	\quad
	\subfigure[]{
		\includegraphics[scale=0.35]{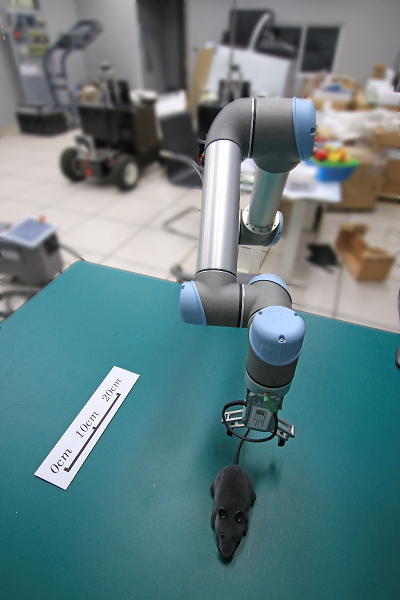}
	}
	\quad
	\subfigure[]{
		\includegraphics[scale=0.35]{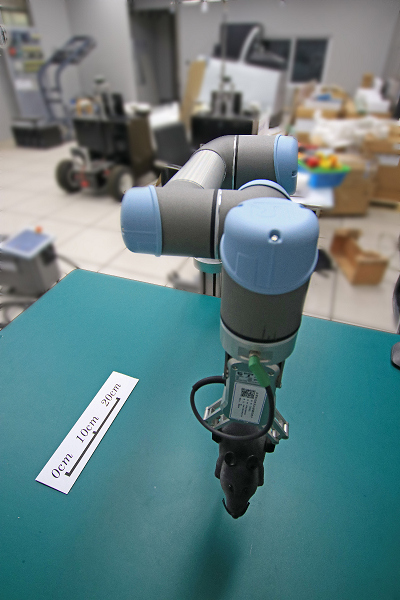}
	}
	\caption{Moving object grasping. 
	(a) The fake mouse is in the initial position. 
	(b) The mouse is moving forward and the robot is changing its control strategy. 
	(c) The mouse keeps moving and the robot keeps tracing it while gripper moving down simultaneously. 
	(d) The robot grasps the mouse successfully.} %图题
	\label{moving-mouse} %标签，方便在文章中引用图片。
\end{figure*}

\subsubsection{Moving Object Grasping}
\label{semantic-tracking}

Grasping a moving object is still a challenging task in visual-based robotic manipulation \cite{moving-grasp}. 
Our learning based approach provides a promising way to approach this challenge. 
To demonstrate the effectiveness of our approach, a case study is conducted in a scenario where a small fake mouse is moving and the robot is ordered to pick up the mouse in motion.

To pick up the moving mouse, our robot needs to be able to track the target continuously and decide to execute a grasping once the action outputted by the control policy is smaller than the preset threshold. 
In order to further reduce the time between making a grasp decision and closing the gripper, 
we add a fixed movement in $z$ direction simultaneously with the movement in the $x-y$ plane, instead of moving down in $z$ direction after the decision making. 
This minor modification significantly improves the successful rate of picking up the mouse in our experiments. 
The example pictures of robot's execution on picking up a moving mouse are shown in Fig.\ref{moving-mouse}. 
However, due to the computation cost of the system and communication delay between the laptop and the robot, the delay time in our current implementation is about $200ms$, which limits the speed of moving objects in our experiments.

\section{Conclusion and future work}
\label{conclusion}
We presented a robotic grasping approach that combines visual perception and a DRL based control policy. 
Comparing with other alternatives, training on a real robot is avoided by decoupling the control from visual perception with the help of a physical representation of objects, which makes them easier to be trained. 
Moreover, the policy trained in simulation could be transferred to a real system without any further training. Real world experiments on UR5 demonstrate the robustness and generalization over a wide variation in challenging grasping tasks.
However, in this work, we only consider 3DOF grasping in which objects are placed on a table and the grasping height is fixed. 
In future work, we would like to extend this work to a 6DOF grasping. 
To do so, it will be important to investigate the pose of gripper in 3D shape perception.

\bibliographystyle{IEEEtran}

\bibliography{refs}

\end{document}